\documentclass[10pt]{article}
\usepackage{hltnaacl04}
\usepackage{times}
\usepackage{xspace}

\usepackage{amssymb}
\usepackage{graphicx}
\usepackage{alltt}

\newcommand{\eg}{\mbox{{e.g.\/},}\xspace}

\newcommand{\name}{content model\xspace}
\newcommand{\nameh}{content-model\xspace}
\newcommand{\names}{content models\xspace}
\newcommand{\naming}{modeling content\xspace}

\newcommand{\Nameshort}{Content}
\newcommand{\Nameh}{Content-model\xspace}
\newcommand{\Names}{Content models\xspace}
\newcommand{\NameS}{Content Models\xspace}

\newcommand{\content}{content structure\xspace}

\newcommand{\ignore}[1]{}
\newcommand{\omt}[1]{}

\newcommand{\state}{s}
\newcommand{\stateindex}{i}
\newcommand{\stateindextwo}{j}
\newcommand{\statelm}[1]{p_{#1}}
\newcommand{\numstates}{m}
\newcommand{\cluster}{c}

\newcommand{\clustone}{\cluster}
\newcommand{\clusttwo}{\cluster'}
\newcommand{\numclusters}{k}
\newcommand{\clustsizethresh}{T}
\newcommand{\insertstate}{\state_\numstates}
\newcommand{\obs}{{\bf x}}
\newcommand{\obslength}{n}
\newcommand{\smoothparam}{\delta}
\newcommand{\smoothemission}{\smoothparam_1}
\newcommand{\smoothtransition}{\smoothparam_2}
\newcommand{\clusterfreq}[1]{f_{#1}}
\newcommand{\clustsentfreq}{D}
\newcommand{\vocab}{V}
\newcommand{\lababbrv}{V-topic\xspace}
\newcommand{\origOrder}{OSO\xspace}
\newcommand{\predAcc}{\origOrder prediction rate\xspace}

\twocolumn

\begin{document}

\title{Catching the Drift: Probabilistic \NameS, with Applications to
Generation and Summarization}

\author{Regina Barzilay \\ Computer Science and AI Lab \\ MIT \\ regina@csail.mit.edu \And Lillian Lee \\ Department of Computer
  Science  \\ Cornell University \\ llee@cs.cornell.edu}
\maketitle

\begin{abstract}

We consider the problem of modeling the {\em content structure} of 
texts within a specific domain, in terms of the topics the texts address and the order in which these
topics appear. We first
present an effective knowledge-lean method for learning \names
from un-annotated  documents, utilizing a novel
adaptation of algorithms for Hidden Markov Models.  We then 
apply our
method to 
two complementary tasks:
information ordering and extractive
summarization.  Our experiments show that incorporating \names
in these applications yields substantial improvement over
previously-proposed methods. 

\medskip
{\bf Publication info:} {\em HLT-NAACL 2004:  Proceedings of the Main Conference}, pp. 113--120.
\end{abstract}

\section{Introduction}
The development and application of computational models of 
text structure is a central concern in natural language processing.
Document-level analysis of text structure is an important instance of
such work.
Previous research has sought
to characterize texts in terms
of domain-independent rhetorical elements, such as schema items
\cite{McKeown:1985a} or rhetorical relations
\cite{Mann&Thompson:1988a,Marcu:1997a}.  The focus of our work,
however, is on an 
equally fundamental but domain-dependent dimension of the
structure of text: {\em
content}.

Our use of the term ``content'' corresponds roughly to the 
notions of
topic and topic change.  
We desire models that can specify, for
example, that articles about earthquakes typically contain information
about 
quake strength, location, and casualties, and that descriptions of
casualties usually precede those of rescue efforts.  But rather than manually
determine the topics for a given domain, we take a {\em
distributional} view, learning them directly from un-annotated texts
via analysis of word distribution patterns.
This idea dates back at least to \newcite{Harris:1982a}, 
who claimed that
``various types of [word] recurrence patterns seem to characterize
various types of discourse''.  Advantages of a distributional perspective
include both drastic reduction in human effort and recognition of
``topics'' that might not occur to a human expert and yet, when
explicitly modeled, 
aid in applications.

Of course, the success of the distributional approach depends on the
existence of recurrent patterns. In arbitrary document
collections, such patterns might be
too variable to be easily detected by statistical means.
However,
research has shown that texts
{\em from the same domain}
tend to exhibit
high
similarity \cite{Wray:2002a}.
Cognitive psychologists have long posited that
this similarity is not accidental,
arguing that formulaic text structure facilitates readers'
comprehension and recall~\cite{Bartlett:1932a}.\footnote{
But ``formulaic'' is not necessarily equivalent to
``simple'', so automated approaches
still offer advantages over manual techniques, especially 
if one needs to model several domains.
}

In this paper, we 
investigate the utility
of domain-specific {\em \names}
for representing topics and topic shifts. 
\Names are
Hidden Markov Models (HMMs) wherein states
correspond to 
types of information characteristic to
the domain of interest  (\eg
earthquake magnitude or previous earthquake occurrences), and state transitions
capture possible information-presentation orderings within that domain.

We first 
describe an efficient, knowledge-lean method for learning both
a set of topics and the relations between topics
directly from un-annotated documents.  Our technique incorporates
a novel adaptation of the standard HMM induction algorithm
that is tailored to the task of \naming.

Then, 
we apply techniques based on \names to
two complex text-processing tasks.  First, we
consider {\em information ordering}, 
that is, choosing a 
sequence in which 
to present
a pre-selected set of items;
this is an essential step in concept-to-text generation, multi-document
summarization, and other text-synthesis problems.  
In our experiments,  \names outperform Lapata's \shortcite{Lapata:2003a} state-of-the-art ordering method
by a wide margin ---
for one domain and performance metric, the gap was 78 percentage points.  
Second,
we consider {\em extractive summarization}: the compression of a
document by choosing a subsequence of its sentences.
For this task, we develop a new \nameh-based learning algorithm for sentence
selection.  The resulting summaries yield 
88\% match with
human-written output, which compares favorably to the 69\% achieved by the
standard ``leading $n$ sentences'' baseline.

The success of \names in these two complementary tasks
demonstrates their flexibility and effectiveness,
and indicates that they are sufficiently expressive to represent
important text properties.  These observations, taken together with
the fact that \names are conceptually intuitive and 
efficiently learnable from raw document collections, suggest
that the formalism  can prove useful in an even broader range of applications
than we have considered here; exploring the options is an appealing
line of future research.

\section{Related Work}

\paragraph{Knowledge-rich methods}

Models employing manual crafting of (typically complex)
representations of content have generally captured  one of 
three types of knowledge \cite{Rambow:1990a,Kittredge&Korelsky&Rambow:1991a}:
{\em domain knowledge} [\eg 
that earthquakes have
magnitudes],
domain-independent {\em  communication knowledge} [\eg 
that  describing an event usually entails specifying its
location]; and {\em domain communication knowledge} [\eg 
that Reuters earthquake reports often conclude by listing previous
quakes\footnote{This does not qualify as domain knowledge because it is not
about earthquakes {per se}.
}].  Form\-alisms exemplifying each of these
knowledge types are DeJong's \shortcite{DeJong:1982a} {\em scripts},
{Mc\-Keown}'s \shortcite{McKeown:1985a} {\em schemas}, and Rambow's
\shortcite{Rambow:1990a} {\em domain-specific schemas}, respectively.

In contrast, because our models are based on a distributional view of content,
they will freely incorporate information from all three categories as long as
such information is manifested as a recurrent pattern.  Also, in comparison to
the formalisms mentioned above, \names constitute a relatively impoverished
representation; but this actually contributes to the ease with which they can
be learned, and our empirical results show that they are quite effective
despite their simplicity.

In recent work, \newcite{Duboue&McKeown:2003a} propose a method for
learning a content planner from a collection of 
texts together with a domain-specific knowledge base,
but our method applies to domains in which no such knowledge base has
been supplied.

\paragraph{Knowledge-lean approaches}  Distributional models of
content have appeared with some frequency in research on {text
segmentation} and topic-based language modeling
~\cite{Hearst:1994a,Beeferman&Berger&Lafferty:1997a,Chen&Seymore&Rosenfeld:1998a,Florian&Yarowsky:1999a,Gildea&Hofmann:1999a,Iyer&Ostendorf:1996a,Wu&Khudanpur:2002a}.
In fact, the methods we employ for learning \names are quite closely
related to techniques proposed in  that literature (see Section
\ref{sec:constr} for more details).  

However, language-modeling research --- whose goal is to predict 
text probabilities --- tends to treat topic as a useful
auxiliary variable rather than a central concern; for example,
topic-based distributional information is generally interpolated with
standard, non-topic-based $n$-gram models to improve probability
estimates.  Our work, 
in contrast, treats content as a
primary entity. In particular, our induction algorithms are designed
with the explicit goal of modeling document content,
which is why they
differ from the standard Baum-Welch (or EM) algorithm for learning
Hidden Markov Models even though \names are instances of HMMs.

\section{Model Construction}
\label{sec:constr}

\label{sec:construction}

We employ an iterative
re-estimation procedure that alternates between (1) creating clusters of text
spans with similar word distributions to serve as representatives of
within-document topics,  and (2) computing models of word
distributions and topic changes
from the clusters so derived.\footnote{For clarity, we omit minor technical details, such as
the use of dummy initial and final states.
Section
\ref{estimation} describes how the free parameters
$\numclusters$, 
$\clustsizethresh$, $\smoothparam_1$, and 
$\smoothparam_2$ are chosen.}

\paragraph{Formalism preliminaries}
We treat texts as sequences of pre-defined text spans, each presumed
to convey information about a single topic.  Specifying text-span
length thus defines the granularity of the induced topics. For
concreteness, in what follows we will refer to ``sentences'' rather
than ``text spans'' since that is what we used in our experiments, but
paragraphs or clauses could 
potentially
have been 
employed
instead.

Our working assumption is that all texts from a given domain  are generated by a single
{\name}. A {\em \name} is an HMM in
which each state $\state$ corresponds to a distinct topic and
generates 
sentences relevant to that topic according to a
state-specific language model $\statelm{\state}$ --- note that 
standard $n$-gram language models can therefore be considered to be
degenerate (single-state) \names.  State transition
probabilities give the probability of changing from a given
topic to another, thereby capturing
constraints on topic shifts.  
We can use the {\em forward algorithm} to 
efficiently compute the generation probability assigned to a document by a \name
and the {\em Viterbi} algorithm to  quickly find the most
likely \nameh state sequence to have generated a given document;
see \newcite{Rabiner:1989a} for details.

In our implementation, we use bigram language models, so that the
probability of an 
$n$-word
sentence $\obs = w_1 w_2 \cdots w_\obslength$ being generated by a
state $\state$ is 
$\statelm{\state}(\obs) \!\stackrel{def}{=}\! \prod_{i=1}^\obslength
\statelm{\state}(w_i|w_{i-1})$.  Estimating the state bigram
probabilities $\statelm{\state}(w'|w)$
is described below.

\begin{figure}[t]
  \centering\small
  \begin{tabular}{|p{.9\columnwidth}|}
    \hline
      The Athens seismological institute said the temblor's epicenter was
      located 380 kilometers (238 miles) south of the capital. \\\hline
      Seismologists in Pakistan's Northwest Frontier Province said the
      temblor's epicenter was about 250 kilometers (155 miles) north of the
      provincial capital Peshawar. \\\hline
      The temblor was centered 60 kilometers (35 miles) northwest of the
      provincial capital of Kunming, about 2,200 kilometers (1,300 miles)
      southwest of Beijing, a bureau seismologist said. \\\hline
\omt{      Beds shook in the provincial capital of Kunming, about 100 kilometers (60
      miles) southeast of the epicenter in Wuding County, said a Yunnan
      seismologist who identified herself only by her surname, Su.\\
    \hline
} 
  \end{tabular}
  \caption{
Samples from  an earthquake-articles sentence cluster,
  corresponding to descriptions of location.}
  \label{fig:cluster}
\end{figure}

\paragraph{Initial topic induction}
As in previous work \cite{Florian&Yarowsky:1999a,Iyer&Ostendorf:1996a,Wu&Khudanpur:2002a},  we initialize the set of ``topics'', distributionally construed,  by
partitioning all of the sentences  from the documents in a given
domain-specific collection into clusters.
{First, we 
create $\numclusters$ clusters via
complete-link clustering, measuring sentence similarity
by
the cosine metric
using word bigrams as features
(Figure~\ref{fig:cluster} shows example output).\footnote{Following \newcite{Barzilay&Lee:2003a}, proper names, numbers and dates are
(temporarily) replaced with generic tokens to help ensure that clusters
contain sentences describing the same event type, rather than same
actual event.}  
Then, given our knowledge that documents may sometimes discuss new and/or irrelevant
content as well, 
we create 
an ``etcetera'' 
cluster
by merging together all clusters containing fewer than
$\clustsizethresh$ sentences, on the assumption that such clusters
consist of
``outlier'' sentences. 
We use $\numstates$ to denote the number
of clusters that results.

\paragraph{Determining states,  emission probabilities,  and
transition probabilities}  
Given a 
set $\cluster_1, \cluster_2, \ldots, \cluster_\numstates$ of
$\numstates$ clusters, where $\cluster_\numstates$ 
is 
the ``etcetera'' cluster,
we construct a \name with corresponding
states $\state_1, \state_2, \ldots, \state_\numstates$; we refer to
$\insertstate$ as the {\em insertion state}.

For each state $\state_\stateindex$, $\stateindex < \numstates$, 
bigram probabilities 
(which  induce the state's sentence-emission probabilities)
are estimated using smoothed
counts from the
corresponding cluster:
$$\statelm{\state_\stateindex}(w'|w) \stackrel{def}{=}  \frac{\clusterfreq{\cluster_\stateindex}(ww')+\smoothemission}{\clusterfreq{\cluster_\stateindex}(w)+\smoothemission|\vocab|},
$$
where $\clusterfreq{\cluster_\stateindex}(y)$ is the frequency with
which word sequence $y$ occurs within the sentences in cluster
$\cluster_\stateindex$, and $\vocab$ is the vocabulary.  But because we want the insertion state
$\insertstate$ to model digressions 
or unseen topics,
we take
the novel step of forcing its language model to be complementary to
those of the other states by setting
 $$\statelm{\insertstate}(w'|w) \stackrel{def}{=}  \frac{1-
 \max_{\stateindex: \stateindex < \numstates} {\statelm{\state_\stateindex}} (w'|w)}{\sum_{u \in \vocab}  (1 -
 \max_{\stateindex: \stateindex < \numstates}
 {\statelm{\state_\stateindex}}(u |w))}.$$ 
Note that the contents of the ``etcetera'' cluster are ignored at this stage.

Our state-transition probability estimates arise from considering how
sentences from the same article are distributed across the clusters.
More specifically, for two clusters $\clustone$ and $\clusttwo$, let
$\clustsentfreq(\clustone,\clusttwo)$ be the number of documents in
which a sentence from $\clustone$ immediately precedes one from
$\clusttwo$, and let $\clustsentfreq(\clustone)$ be the number of documents
containing sentences from $\cluster$.  Then, for any two states
$\state_\stateindex$ and $\state_\stateindextwo$, $\stateindex,
\stateindextwo \leq \numstates$, we use the following smoothed
estimate of the probability of transitioning from $\state_\stateindex$
to $\state_\stateindextwo$: 
$$p(\state_\stateindextwo|\state_\stateindex) = \frac
{\clustsentfreq(\cluster_\stateindex,\cluster_\stateindextwo) + \smoothtransition}
{\clustsentfreq(\cluster_\stateindex) + \smoothtransition\numstates}.$$

\paragraph{Viterbi re-estimation} 

Our initial clustering ignores
sentence order; however, contextual
clues may indicate that sentences with high lexical similarity 
are actually on different ``topics''.
For instance, Reuters articles about
earthquakes frequently finish by mentioning previous quakes.  This
means that while the sentence 
``The temblor injured dozens'' at the beginning of a
report is probably highly salient and should
be included in a summary of it,  the same sentence at the
end of the piece probably refers to a different event, and so should be omitted.

A natural way to incorporate ordering information is iterative
re-estimation of the model parameters, since the \name itself provides
such information through its transition structure. We take an EM-like {\em
Viterbi} approach \cite{Iyer&Ostendorf:1996a}: we re-cluster
the sentences by placing each one in the (new) cluster
$\cluster_{\stateindex}$, $\stateindex \leq \numstates$, that corresponds
to the state $\state_{\stateindex}$  most likely to have
generated it according to the Viterbi decoding of the training data.
We then use this new clustering as the input to the
procedure for estimating HMM parameters described above.
The cluster/estimate
cycle is repeated until the clusterings stabilize or we reach  a
predefined number of iterations.

\section{Evaluation Tasks}

We apply the techniques just described to two tasks 
that stand to benefit from models of content and 
changes in topic:
information ordering for text generation and information selection for
single-document summarization.  These are two complementary tasks that rely on
disjoint model functionalities: the ability to order a set of
pre-selected information-bearing items, and the ability
to do the selection itself, extracting from an ordered sequence of information-bearing items a
representative subsequence.

\subsection{Information Ordering}

The information-ordering task
is 
essential 
to
many text-synthesis 
applications, 
including concept-to-text generation and multi-document
summarization;
While accounting for the full 
range of discourse and stylistic
factors that influence the ordering process
is
infeasible in
many domains, 
 probabilistic \names provide a
means for
handling important aspects of this 
problem.
We demonstrate this point by utilizing \names to select appropriate
sentence orderings: we simply use a \name trained on documents from
the domain of interest, selecting the ordering among all the presented
candidates that the \name assigns
the highest probability to.

\subsection{Extractive Summarization}

\Names can also be used for
single-document summarization.  Because ordering is not an issue in this
application\footnote{Typically, sentences in a single-document summary follow
  the order of appearance in the original document.}, this task tests the
ability of \names to adequately represent domain topics independently of
whether they do well at ordering these topics.

The usual strategy employed by domain-specific summarizers is for
humans to determine
\emph{a priori} what types of information from the originating
documents
should be included 
(\eg in stories about earthquakes, the number of victims)
\cite{Radev&McKeown:1998a,Whiteetal:2001a}.
Some 
systems
avoid the need for manual analysis by learning content-selection rules from a
collection of articles paired with human-authored
summaries,  
but  their learning algorithms typically 
focus on within-sentence features or very coarse structural features
(such as position within a paragraph)~\cite{Kupiec&Pedersen&Chen:1999a}.
Our \nameh-based summarization algorithm
 combines the advantages
of both approaches; on the one hand, it learns all required information
from 
un-annotated document-summary pairs; on the
other hand, it operates on a more abstract and global level,
making use of the topical structure of the entire document.  

Our algorithm is trained as follows.
Given a \name acquired from the full articles
using the method described in Section \ref{sec:construction}, 
we need to
learn which
topics
(represented by the \name's states) should
appear in our summaries.
Our first step is to employ the Viterbi algorithm to
tag all of the summary sentences and all of the sentences from the original articles with
a {\em Viterbi topic label}, or {\em \lababbrv}  --- the name of the state most likely to have
generated them.  Next, 
 for each state $\state$  such that at least three full
training-set articles contained \lababbrv $s$, we 
compute the probability that the state generates sentences
that should appear in a summary.  This probability is  estimated by simply 
(1) counting the number of document-summary pairs in the parallel training data
such that both the originating document and the summary contain
sentences assigned \lababbrv $\state$, and then (2) normalizing this count by 
the number of full articles  containing sentences with  \lababbrv $\state$.

To produce a length-$\ell$ summary of a new article, the algorithm first uses
the \name and Viterbi decoding to assign each of the article's sentences a
\lababbrv. Next, the algorithm selects those $\ell$ states, chosen from among
those that appear as the \lababbrv of one of the article's sentences, that have
the highest probability of generating a summary sentence, as estimated above.
Sentences from the input article corresponding to these states are placed in
the output summary.\footnote{If there are more than $\ell$ sentences, we
  prioritize them by the summarization probability of their \lababbrv's state;
  we break any further ties by order of appearance in the document.}

\section{Evaluation Experiments}

\subsection{Data}

For evaluation purposes, we created corpora from five domains:
earthquakes, clashes between armies and rebel groups, drug-related criminal offenses,
financial reports, and summaries of aviation
accidents.\footnote{\scriptsize
  \texttt{http://www.sls.csail.mit.edu/\char`\~regina/struct}} Specifically,
the first four collections 
consist of AP articles from the North American News Corpus gathered
via a TDT-style document clustering system.  The fifth 
consists of narratives from the National Transportation Safety Board's
database  previously employed by \newcite{Jones&Thompson:2003a} for 
event-identification experiments.
For each such set, 100 articles
were used for training a \name, 100 articles for testing, and 20 for the
development set used for parameter tuning. 
\begin{table}[t]
  \centering
  \begin{tabular}{|@{~}l@{~}|@{~}c@{~}|@{~}c@{~}|@{~}c@{~}|@{~}r@{~}|}
    \hline
    Domain     & Average
               & Standard
               & Vocabulary
               & \multicolumn{1}{@{}c@{~}|}{Token/} \\
               & Length
               & Deviation
               & Size
               & \multicolumn{1}{@{}c@{~}|}{type} \\\hline
    Earthquakes & 10.4  & 5.2  & 1182       & 13.2 \omt{13.158} \\\hline
    Clashes    &  14.0  & 2.6  & 1302       & 4.5 \omt{4.464} \\\hline
    Drugs      &  10.3  & 7.5  & 1566       & 4.1 \omt{4.098} \\\hline
    Finance    &  13.7  & 1.6  & 1378       & 12.8 \omt{12.821} \\\hline
    Accidents  &  11.5  & 6.3  & 2003       & 5.6 \omt{5.556} \\\hline
  \end{tabular}
  \caption{Corpus statistics. 
  Length is in sentences.  Vocabulary size and
  type/token ratio are computed after replacement of proper names, numbers
  and dates.}
  \label{table:domain-statistics}
\end{table}
Table~\ref{table:domain-statistics} presents information about article length
(measured in sentences, as determined by 
the  sentence
separator of ~\newcite{Reynar&Ratnaparkhi:1997a}), vocabulary size,
and token/type ratio  for
each domain.

\subsection{Parameter Estimation}
\label{estimation}

Our training algorithm has four free parameters: 
two that indirectly control the number of states in the induced
\name, and two parameters for smoothing bigram probabilities.
All 
 were tuned separately for each domain on the
corresponding held-out development set using Powell's grid
search~\cite{Press&Teukolsky&Vetterling&Flannery:1997a}.  The
parameter values were selected to optimize 
system performance on
the information-ordering task\footnote{See Section~\ref{sum+order} for
discussion of the relation between the ordering and the
summarization task.}.
We found that across all domains, the optimal models 
were based on ``sharper''
language models 
(e.g., $\smoothemission< 0.0000001$).
The optimal number of states 
ranged
from 32 to 95.

\subsection{Ordering Experiments}
\label{order-eval}

\subsubsection{Metrics} The intent behind our {\em ordering}
experiments is to test whether \names assign high probability to
acceptable sentence arrangements.  However, one stumbling block to
performing this kind of evaluation is that we do not have data on ordering
quality: the set of 
sentences from an $N$-sentence document can be 
sequenced in $N!$ different ways, which
even for a single text of moderate length is too many to ask humans to
evaluate.  Fortunately, we do know that at least the
original sentence order ({\em \origOrder}) in the source document must
be acceptable, and so we should prefer algorithms that assign it high
probability relative to the bulk of all the other possible
permutations.
This observation motivates our first evaluation metric:
the {\em rank} received by the \origOrder when all
permutations of a given document's sentences are sorted by the
probabilities that the model under consideration assigns to them.  The
best possible rank is 0, and the worst is $N! -1$.

An additional difficulty we encountered in setting up our evaluation
is that while we wanted to compare our algorithms against Lapata's
\shortcite{Lapata:2003a} state-of-the-art system, her method
doesn't consider all permutations (see below), and so the rank metric cannot
be computed for it.  To compensate, we report the {\em \predAcc},
which measures the percentage of test cases in which the model under
consideration gives highest probability to the \origOrder from among all
possible permutations; we expect that a good model should predict the
\origOrder a fair fraction of the time.  Furthermore, to provide
some assessment of the quality of the predicted orderings themselves,
we follow \newcite{Lapata:2003a} in employing {\em Kendall's $\tau$},
which
is a measure of how much an ordering differs from the \origOrder --- 
the underlying assumption is  that most reasonable sentence orderings should be
fairly similar to it.  
Specifically, for a  permutation $\sigma$ of the sentences in an
$N$-sentence document,  $\tau(\sigma)$ is computed as
$$\tau(\sigma) = 1 - 2\frac{S(\sigma)}{{N \choose 2}},$$
where 
$S(\sigma)$ is the number of swaps of adjacent sentences necessary to
re-arrange $\sigma$  into the \origOrder. The metric ranges from -1 (inverse orders)
to 1 (identical orders). 

\subsubsection{Results}

For each of the 500 unseen test texts, we exhaustively enumerated all
sentence permutations and ranked them using a \name from the
corresponding domain.  We compared our results against those of a
bigram language model (the baseline) and an improved version of the
state-of-the-art probabilistic ordering method of
Lapata~\shortcite{Lapata:2003a}, both trained on the same data we
used.  Lapata's method first learns a set of pairwise
sentence-ordering preferences based on features such as noun-verb
dependencies.  Given a new set of sentences, the latest version of her
method applies a
Viterbi-style approximation algorithm to choose a permutation
satisfying many preferences (Lapata, personal communication).\footnote{Finding the optimal such permutation is
NP-complete.}

Table~\ref{table:ordering} gives the results of our ordering-test
comparison experiments.  
\Names outperform the alternatives almost universally, and often by a
very wide margin. We conjecture that this difference in performance
stems from
the ability of \names to capture  \emph{global} document structure.
In contrast, the other two algorithms are local, taking into account only
the relationships between adjacent word
pairs and adjacent sentence pairs, 
respectively.  It is interesting to observe that our method achieves better results
despite not having access to the linguistic information incorporated
by Lapata's method.  To be fair, though, her techniques were designed
for a larger corpus than ours, which may aggravate data sparseness
problems for such a feature-rich method.

Table~\ref{table:ordering-rank} gives further details on the rank
results for our \names, showing how the rank scores were distributed;
for instance, we see that on the Earthquakes domain,  the \origOrder
was one of the top five permutations in 95\% of the
test documents.
Even in Drugs and Accidents --- the domains that proved relatively
challenging to our method --- in more than 55\% of the
cases the \origOrder's rank did not exceed ten.  Given that the maximal
possible rank in these domains 
exceeds three million, we believe that our model has done a good job in the
ordering task.

We also computed learning curves for the different domains; these are shown in
Figure~\ref{fig:learn-curve}.  Not surprisingly, performance improves with the
size of the training set for all domains.  The figure also shows that the
relative difficulty (from the \nameh point of view) of the different domains
remains mostly constant across varying training-set sizes.  Interestingly, the
two easiest domains, Finance and Earthquakes, can be thought of as being more
formulaic or at least more redundant, in that they have the highest token/type
ratios (see Table~\ref{table:domain-statistics}) --- that is, in these domains,
words are repeated much more frequently on average.

\begin{table}[t]
  \centering
{ 
\newcommand{\B}[1]{{\bf #1}}
\newcommand{\notavail}{(N/A)}
  \begin{tabular}{|l|l|r|r|r|} \hline
    {Domain}    & System  & \multicolumn{1}{|c|}{Rank}      &\origOrder & \multicolumn{1}{|c|}{$\tau$} \\
              &            &           & \multicolumn{1}{|c|}{pred.}&          \\
\hline\hline
              & \Nameshort & \B{2.67}  & \B{72\%} & \B{0.81} \\\cline{2-5}
    Earthquakes& Lapata  & \notavail &  24\%    & 0.48     \\\cline{2-5}
              & Bigram     & 485.16    &   4\%    & 0.27     \\\hline\hline
              & \Nameshort & \B{3.05}  & \B{48\%} & \B{0.64} \\\cline{2-5}
    Clashes   & Lapata  & \notavail &  27\%    & 0.41     \\\cline{2-5}
              & Bigram     & 635.15    &  12\%    & 0.25     \\\hline\hline
              & \Nameshort & \B{15.38} & \B{38\%} & {0.45}   \\\cline{2-5}
    Drugs     & Lapata  & \notavail &  27\%    & \B{0.49} \\\cline{2-5}
              & Bigram     & 712.03    &  11\%    & 0.24     \\\hline\hline
              & \Nameshort & \B{0.05}  & \B{96\%} & \B{0.98} \\\cline{2-5}
    Finance   & Lapata  & \notavail &  18\%    & 0.75     \\\cline{2-5}
              & Bigram     & 7.44      &  66\%    & 0.74     \\\hline\hline
              & \Nameshort & \B{10.96} & \B{41\%} & \B{0.44} \\\cline{2-5}
    Accidents & Lapata  & \notavail &  10\%    & 0.07     \\\cline{2-5}
              & Bigram     & 973.75    &   2\%    & 0.19     \\\hline
  \end{tabular}
} 
  \caption{Ordering results (averages over the test cases).}
  \label{table:ordering}
\end{table}

\begin{table}[t]
  \centering
  \begin{tabular}{|l|r|c|c|}
    \hline
    Domain             & \multicolumn{3}{c|}{Rank range}  \\\cline{2-4}
                       & \multicolumn{1}{|c|}{[0-4]}      & [5-10]    & $>10$  \\\hline 
    Earthquakes         &  95\%       & 1\%       & 4\%   \\\hline
    Clashes            &  75\%       & 18\%      & 7\%   \\\hline
    Drugs              &  47\%       & 8\%       & 45\%  \\\hline
    Finance            &  100\%      & 0\%       & 0\%   \\\hline
    Accidents          &  52\%       & 7\%       & 41\%  \\\hline
    
  \end{tabular}
  \caption{Percentage of cases for which the \name assigned to the \origOrder a rank within a
  given range.}
  \label{table:ordering-rank}
\end{table}

\begin{figure}[t]
  \fbox{\includegraphics[scale=0.3,angle=270]{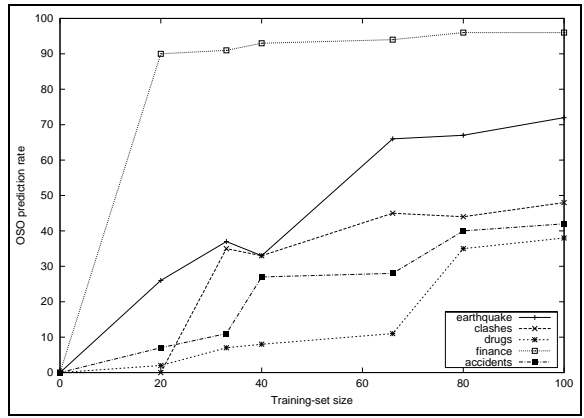}}
  \caption{
Ordering-task performance, in terms of \predAcc, as a function of the number of
documents in the training set.}
  \label{fig:learn-curve}
\end{figure}

\subsection{Summarization Experiments}
\label{summar-eval}

The evaluation of our summarization algorithm
was driven by two questions: (1) Are the summaries produced of acceptable
quality, in terms of selected content?  and (2) Does the \nameh representation
provide additional advantages over 
more locally-focused methods?

To address the first question, we compare
summaries created by our system against the ``lead'' baseline, which 
extracts the  first $\ell$ sentences of the original text ---
despite its
simplicity, the results from the annual Document Understanding Conference (DUC)
evaluation 
suggest that most single-document summarization systems cannot beat this
baseline. To address  question (2), we consider a summarization system
that learns extraction rules directly from a parallel corpus of full texts and
their summaries~\cite{Kupiec&Pedersen&Chen:1999a}. In this system,
summarization is framed as a sentence-level binary classification
problem: each sentence is labeled by the publicly-available
{BoosTexter} system~\cite{Schapire&Singer:2000a} as being either ``in'' or ``out'' of the summary.
The
features considered for each sentence are its unigrams and 
its location within the text, namely
\emph{beginning third}, \emph{middle third} and \emph{end
  third}.\footnote{This feature set yielded the best results among the
  several possibilities we tried.
} Hence, relationships between sentences are not explicitly
modeled, making this system a good basis for comparison.

We evaluated our summarization system on 
the Earthquakes domain, since for
some of the texts in this domain there is a condensed version written by AP
journalists.
These summaries are mostly extractive\footnote{Occasionally, one or
  two phrases or, more rarely, a clause
  were dropped.}; consequently, they can be
easily aligned with sentences in the original articles. From sixty
document-summary pairs,
half were randomly selected to be used for training and the other half for
testing. (While thirty documents may not seem like a large number, it
is comparable to the size of the training
corpora used in the competitive summarization-system evaluations mentioned
above.)
The average number of sentences in the full texts and summaries was 15 and 6,
respectively, for a total of
450 sentences in each of the test and (full documents of the) training sets.

At runtime, we provided the systems with a full document and the
desired output length, namely, the length in sentences of the
corresponding shortened version.  The resulting summaries were judged
as a whole
by the fraction of their component sentences that appeared in the
human-written summary of the input text.

The results in Table~\ref{table:sum-results} confirm our hypothesis about the
benefits of \names for text summarization --- our model outperforms
both the sentence-level, locally-focused 
classifier 
and the ``lead'' baseline.  Furthermore,
as the learning curves shown in Figure~\ref{fig:sum-curve} indicate,
our method
achieves good performance on a small subset of parallel training data:
in fact, the accuracy of our method on 
one third of the
training data is higher than that of the sentence-level classifier on the full
training set. Clearly, this performance gain demonstrates the effectiveness of
\names for the summarization task.

\begin{table}[t]
  \begin{center}
    \begin{tabular}{|l|c|c|c|}
      \hline      
System          & Extraction accuracy   
      \\\hline 
      {\Nameshort-based}    & \textbf{88\%} \\\hline
      Sentence classifier   & 76\%          \\
      (words + location)    &               \\\hline
      Leading $n$ sentences & 69\%          \\\hline
    \end{tabular}
  \end{center}
  \caption{
    Summarization-task results.}
  \label{table:sum-results}
\end{table}

\begin{figure}[t]
  \fbox{\includegraphics[scale=0.6]{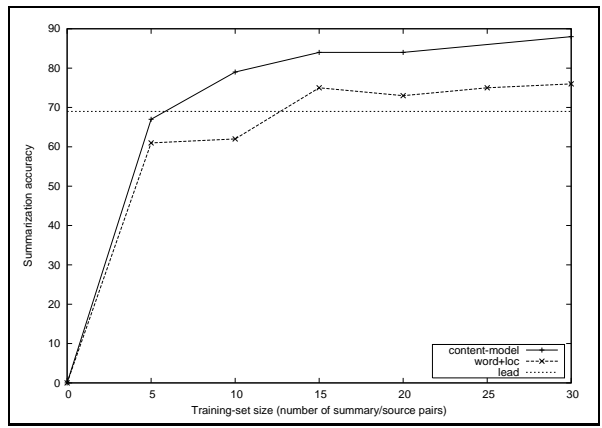}}
  \caption{
Summarization performance (extraction accuracy) on Earthquakes as a function of  training-set size.}
  \label{fig:sum-curve}
\end{figure}

\subsection{Relation Between Ordering and Summarization Methods}\label{sum+order}

\begin{table}[t]
  \centering
\begin{tabular}{|l@{~}|*{5}{@{~}c@{~}|}c|}    \hline
    Model size       & 10     &  20      & 40    & 60   & {\bf 64} &
  80    \\\hline 
    Ordering        & 11\%   &  28\%    & 52\%  & 50\% & {\bf 72\%} & 57\%    \\\hline
    Summarization    & 54\%   &  70\%    & 79\%  & 79\% & {\bf 88\%} & 83\%   \\\hline
    
  \end{tabular}
  \caption{\Nameh performance  on Earthquakes     
as a function of model size.  Ordering: \predAcc;
  Summarization: extraction accuracy.}
  \label{table:order+sum}
\end{table}

Since we used two somewhat orthogonal tasks,  ordering and
summarization, to evaluate the
quality of the \nameh paradigm, it is interesting to ask whether the same 
parameterization of the model does well in both cases. Specifically,
we looked at the results for different model topologies, induced by
varying the number of \nameh states.
For these tests, we experimented with the Earthquakes data (the
only domain for which we could evaluate summarization performance), and exerted direct control over the number of states, rather than
utilizing the cluster-size threshold; that is, in order to create
exactly $m$ states for a specific value of $m$, we merged the smallest
clusters until $m$ clusters remained.

Table~\ref{table:order+sum} shows the performance of the
different-sized \names with respect to the summarization task and the
ordering task (using \predAcc). While the ordering 
results
seem to be
more sensitive to the number of states, both metrics induce similar
ranking on the models. In fact, the same-size model yields top performance
on both tasks. While our experiments are limited to only one domain,
the correlation in results is encouraging: optimizing parameters on
one task promises 
to yield good performance on the other.
These findings provide support for the hypothesis that \names are not
only helpful for specific tasks, but can serve as effective
representations of text structure in general.

\section{Conclusions}

In this paper, we present an unsupervised method for the induction of \names,
which capture constraints on topic selection and organization for texts in a
particular domain.  Incorporation of these models in ordering and summarization
applications yields substantial improvement over previously-proposed methods.
These results indicate that distributional approaches widely used to model
various inter-sentential phenomena can be successfully applied to capture
text-level relations, empirically validating the long-standing hypothesis that
word distribution patterns strongly correlate with discourse patterns within a
text, at least within specific domains.

An important future direction lies in studying the correspondence
between our domain-specific model and domain-independent formalisms,
such as RST.  By automatically annotating a large corpus of texts with
discourse relations via a rhetorical 
parser
\cite{Marcu:1997a,Soricut&Marcu:2003a},
we 
may be able to  
incorporate domain-independent relationships into the transition
structure of our \names.
This study 
could 
uncover
interesting connections between domain-specific stylistic
constraints and generic principles of text organization.

In the literature, discourse is frequently modeled using a hierarchical
structure, which suggests that probabilistic context-free grammars or
hierarchical Hidden Markov Models \cite{Fine+Singer+Tishby:98a} may
also be applied for modeling \content.  
In the future, we
plan to investigate how to bootstrap the induction of 
hierarchical models
using labeled data derived from 
our \names. We 
would also like to
explore how domain-independent discourse constraints can be used to
guide the construction of the hierarchical models.

{
{
\paragraph{Acknowledgments} 
We are grateful to Mirella Lapata for providing us the results of her system on
our data, and to Dominic Jones and Cindi Thompson for supplying us with their
document collection. We also thank Eli Barzilay, Sasha Blair-Goldensohn, Eric
Breck, Claire Cardie, Yejin Choi, Marcia Davidson, Pablo Duboue, No\'emie
Elhadad, Luis Gravano, Julia Hirschberg, Sanjeev Khudanpur, Jon Kleinberg, Oren
Kurland, Kathy McKeown, Daniel Marcu, Art Munson, Smaranda Muresan, Vincent Ng,
Bo Pang, Becky Passoneau, Owen Rambow, Ves Stoyanov, Chao Wang and the
anonymous reviewers for helpful comments and conversations.  Portions of this
work were done while the first author was a postdoctoral fellow at Cornell
University.  This paper is based upon work supported in part by the National
Science Foundation under grants ITR/IM IIS-0081334 and IIS-0329064 and by an
Alfred P.  Sloan Research Fellowship. Any opinions, findings, and conclusions
or recommendations expressed above are those of the authors and do not
necessarily reflect the views of the National Science Foundation or Sloan
Foundation.  } }

\newcommand{\bibsnip}{\vspace*{-.1in}}
\bibliographystyle{scrunchacl}
{

}
\end{document}